\documentclass{article}

\PassOptionsToPackage{numbers, compress}{natbib}

 \usepackage[dblblindworkshop, final]{neurips_2025}
\workshoptitle{AI for Science}

\usepackage[utf8]{inputenc} 
\usepackage[T1]{fontenc}    
\usepackage{hyperref}       
\usepackage{url}            
\usepackage{booktabs}       
\usepackage{amsfonts}       
\usepackage{nicefrac}       
\usepackage{microtype}      
\usepackage{xcolor}         
\usepackage[inline]{enumitem}
\usepackage[nolist,nohyperlinks]{acronym}
\usepackage{amsmath}
\usepackage{graphicx}
\usepackage{multirow}
\usepackage{makecell}
\usepackage[capitalize,noabbrev]{cleveref}
\usepackage{subcaption} 
\usepackage[normalem]{ulem}
\usepackage{siunitx}

\usepackage{svg}
\usepackage{subcaption}
\usepackage{geometry}
\usepackage{wrapfig}
\usepackage{tikz}

\newcommand{\textbfp}[1]{\vspace{0.5em}\noindent\textbf{#1.}}

\title{Towards Multi-Fidelity Scaling Laws \\ of Neural Surrogates in CFD}

\begin{acronym}[LONGEST]
  \acro{LLM}{Large Language Model}
  \acro{CV}{Computer Vision}
  \acro{PDE}{Partial Differential Equation}
  \acro{CFD}{Computational Fluid Dynamics}
  \acro{LES}{Large Eddy Simulation}
  \acro{RANS}{Reynolds-Averaged Navier-Stokes}
  \acro{Re}{Reynolds number}
  \acro{AoA}{Angle of Attack}
  \acro{FVM}{Finite Volume Method}
  \acro{DNS}{Direct Numerical Simulation}
  \acro{MSE}{Mean Squared Error}
  \acro{SOTA}{state of the art}
  \acro{nMAE}{normalized Mean Absolute Error}

\end{acronym}

\author{
  \textbf{Paul Setinek}$\:^{1}$ \hspace{1em}
  \textbf{Gianluca Galletti}$\:^{1}$ \hspace{1em}
  \textbf{Johannes Brandstetter}$^{1,2}$
  \vspace{4px}\\
  {$^1$~ELLIS Unit, LIT AI Lab, Institute for Machine Learning, JKU Linz, Austria}\\
  {$^2$~Emmi AI, Linz, Austria}\\\vspace{2px}
  \texttt{setinek@ml.jku.at}
}

\begin{document}

\maketitle

\begin{abstract}
Scaling laws describe how model performance grows with data, parameters and compute.
While large datasets can usually be collected at relatively low cost in domains such as language or vision, scientific machine learning is often limited by the high expense of generating training data through numerical simulations.
However, by adjusting modeling assumptions and approximations, simulation fidelity can be traded for computational cost, an aspect absent in other domains.
We investigate this trade-off between data fidelity and cost in neural surrogates using low- and high-fidelity Reynolds-Averaged Navier-Stokes (RANS) simulations.
Reformulating classical scaling laws, we decompose the dataset axis into compute budget and dataset composition.
Our experiments reveal compute-performance scaling behavior and exhibit budget-dependent optimal fidelity mixes for the given dataset configuration.
These findings provide the first study of empirical scaling laws for multi-fidelity neural surrogate datasets and offer practical considerations for compute-efficient dataset generation in scientific machine learning.
\end{abstract}

\section{Introduction}
\label{sec:introduction}

Machine learning has seen immense progress in recent years, which was not only driven by architectural or methodological innovations but also by the increasing availability of computational resources.
This has enabled the scaling up of models, and as a result many SOTA models now contain hundreds of billions of parameters \citep{grattafiori2024llama3herdmodels, mistral2025large2}.
Scaling laws, which originated in the domain of \acp{LLM} \citep{kaplan2020scaling, hoffman2022chinchilla}, have expanded into various other areas like \ac{CV} \citep{zhai2022scaling_vit} or time series \citep{shi2024scaling_timeseries, yao2025scaling_timeseries}.
These empirical studies describe how models improve as a function of three \textit{axes}:
\begin{enumerate*}[label=(\roman*)]
    \item parameter count ($N$),
    \item dataset size ($D$), and
    \item compute ($C$).
\end{enumerate*}

In the meantime, scientific machine learning has similarly achieved remarkable success in modeling complex systems with neural surrogates.
Notable examples include breakthroughs in weather and climate forecasting \citep{keisler2022forecasting_global_weather, pathak2022fourcastnet, nguyen2023climax, sanchez2025gencast, bodnar2024aurora}, material design \citep{merchant2023scaling_dl_for_materials, zeni2025generative_materials, yang2024matter_sim} or protein folding \citep{jumper2021alphafold, abramson2024alphafold3}.
These advances have partly been enabled by large curated public datasets, such as ERA5 in weather modeling \citep{hersbach2023era5} or the Protein Data Bank (PDB) \citep{berman2000pdb}.
More recently, first large scale datasets have also been released in areas such as automotive aerodynamics \citep{ashton2024drivearml, elrefaie2024drivearnetpp}.

However, in many areas of science and engineering, such datasets are either not public or do not exist and therefore researchers are required to generate their own problem specific dataset prior to model training.
Since the systems of interest are usually governed by \acp{PDE} \citep{evans2010pde}, generating training data requires solving these equations numerically, which is often coupled with significant computational costs \citep{yang2013simulation}.
This means that unlike other domains, where data can be sourced with little to no computation (\textit{e.g.} text, real-world images and videos), the dataset size $D$ in scientific machine learning is often \uline{no longer ``free'' to scale}.
While prior works study how performance scales with dataset size, they do not take the computational cost associated with scaling the dataset axis into account \citep{subramanian2023towards_foundation_models, herde2024poseidon, alkin2024upt, paischer2025gyroswin}.
Moreover, many neural surrogates exist with the goal of ``amortizing'' the training and dataset cost with repeated, cheap evaluations down the line; if dataset generation becomes prohibitively expensive, it fundamentally defeats the purpose of such models.

The numerical solution of \acp{PDE} is a long-running research topic, which can be very nuanced: when designing numerical simulations, modeling assumptions and simplifications of the underlying physics need to be made, which trade fidelity for compute.
Ordered by computational complexity, common approaches include \ac{DNS} which resolves all turbulent scales at prohibitive cost; \ac{LES}, which resolves only large scales while modeling subgrid-scale dynamics; and \ac{RANS}, where turbulence is entirely modeled through statistical averaging, making \ac{RANS} the cheapest but also least accurate of the three.
For example, a 3D \ac{LES} over an airfoil can take several orders of magnitude longer than simulating the respective \ac{RANS} equations.
Even on simple scenarios, \ac{LES} take an order of magnitude longer than \ac{RANS} simulations \cite{Lopes2024lesvrrans}, a factor which grows substantially for more chaotic systems.

This introduces a fundamental trade-off, which motivates our research question:

\begin{quote}
    \textit{Under a fixed budget constraint, what is the optimal training set composition of low- and high-fidelity samples in order to maximize model performance?}
\end{quote}

We move towards an answer to this essential question by proposing a reformulation of classical scaling laws, devised to account for this phenomenon.
We argue for splitting the simple \textit{dataset size} $D$ axis into two components, namely \uline{\emph{dataset compute budget $D_b$}} (core hours allocated for data generation) and \uline{\emph{dataset composition $D_c$}} (controls the fidelity distribution of the training data).

Since training data generation is typically the primary bottleneck in scientific machine learning, we focus on these two axes while assuming model size and compute for model training are non-limiting.
Although restrictive, this assumption allows us to directly address our research question that has not been studied in prior work.
Our contributions can be summarized as follows:
\begin{itemize}
    \item We introduce a formulation of multi-fidelity scaling laws, extending classical scaling law analysis to settings where training data can be simulated at different fidelities.
    \item We design a multi-fidelity dataset of external aerodynamics around airfoils, incorporating different modeling assumptions across fidelity levels.
    \item We present the first empirical investigation of multi-fidelity scaling laws along the data axis on a \ac{CFD} dataset, evaluating how model performance scales under varying dataset compositions and generation budgets.
\end{itemize}

\section{Related work}
\label{sec:related_work}

\textbfp{Learning from multi-fidelity data}
Multi-fidelity data exists in different fields, and can come in different shapes and forms (e.g., varying realism, accuracy or resolution). In \ac{CV}, \textit{``resolution transfer''} \citep{beyer2023flexivitmodelpatchsizes, schusterbauer2024boosting} or \textit{``super-resolution''} \citep{dong2015imagesuperresolutionusingdeep, Moser_2025} are well researched directions, where models learn to predict fine-scale details from coarse observations. 
Similar ideas appear in scientific machine learning under \textit{``discretization convergence''} in neural operators \citep{kovachki2021neural}, where models are trained to generalize across mesh resolutions. While these methods may be invariant to changes in resolution (even though most of them have no theoretical guarantee \citep{bartolucci2023representationequivalentneuraloperators}), they do not capture underlying physics and fidelity shifts.

\textbfp{Transfer learning for multi-fidelity data in scientific machine learning}
Recent studies have explored transfer learning from low- to high-fidelity data \citep{de2020transfer_learning_mf, song2021transfer_learning_mf, lu2022multifidelityneuraloperators}.
However, in these works the distinction between fidelities is limited to changes in mesh resolution, while the underlying physical modeling assumptions remain the same.
In contrast, a more recent study takes this further by transferring knowledge from low-fidelity \ac{RANS} to high-fidelity \ac{LES} simulations in the context of wind farm modeling, thereby altering the underlying modeling assumptions \citep{zhang2024windfarm}.
While these studies address an important aspect of scientific machine learning, our investigation pursues a complementary goal: we focus on identifying patterns suggesting the existence of optimal strategies for generating training data to maximize the performance of neural surrogates.

\section{Dataset}
\label{sec:data_generation}

\begin{figure}[t]
    \centering
    
    \begin{subfigure}[b]{0.85\textwidth}
        \centering
        \includegraphics[width=\textwidth]{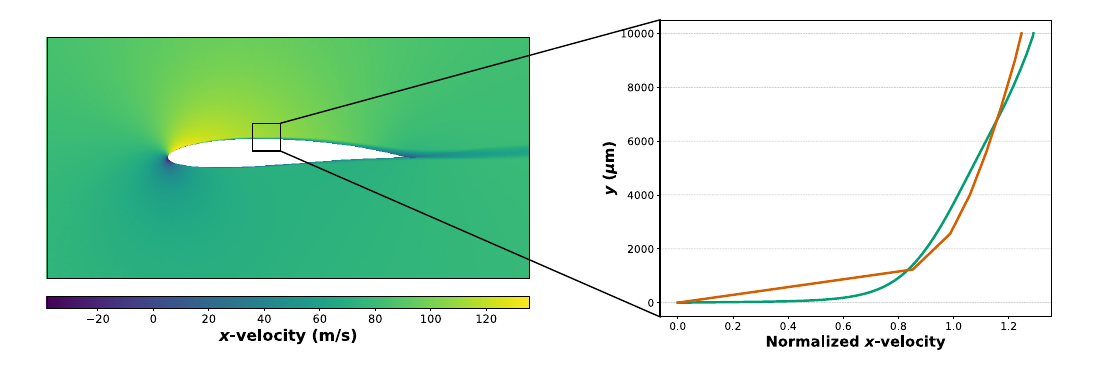}
        \caption{Velocity field around the airfoil (left) and boundary layer detail (right).}
        \label{fig:af_sub1}
    \end{subfigure}

    \vspace{0.2em}

    \begin{subfigure}[b]{0.3\textwidth}
        \centering
        \includegraphics[width=\linewidth]{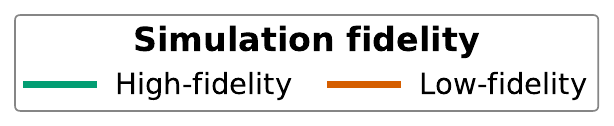}
    \end{subfigure}

    \vspace{0.3em}
    \begin{subfigure}[b]{0.4\textwidth}
        \centering
        \includegraphics[width=\textwidth]{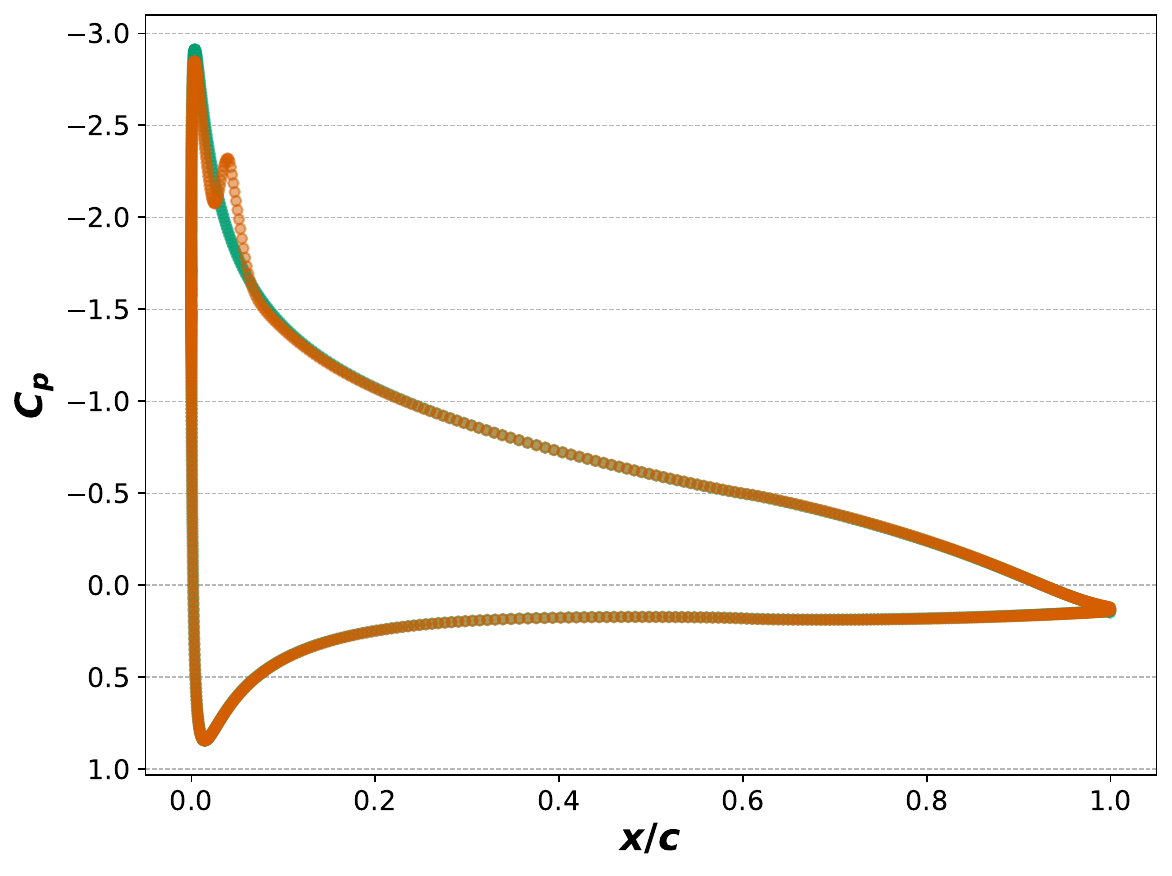}
        \caption{Pressure Distribution along the chord.}
        \label{fig:af_sub2}
    \end{subfigure}
    \hspace{24px}
    \begin{subfigure}[b]{0.4\textwidth}
        \centering
        \includegraphics[width=\textwidth]{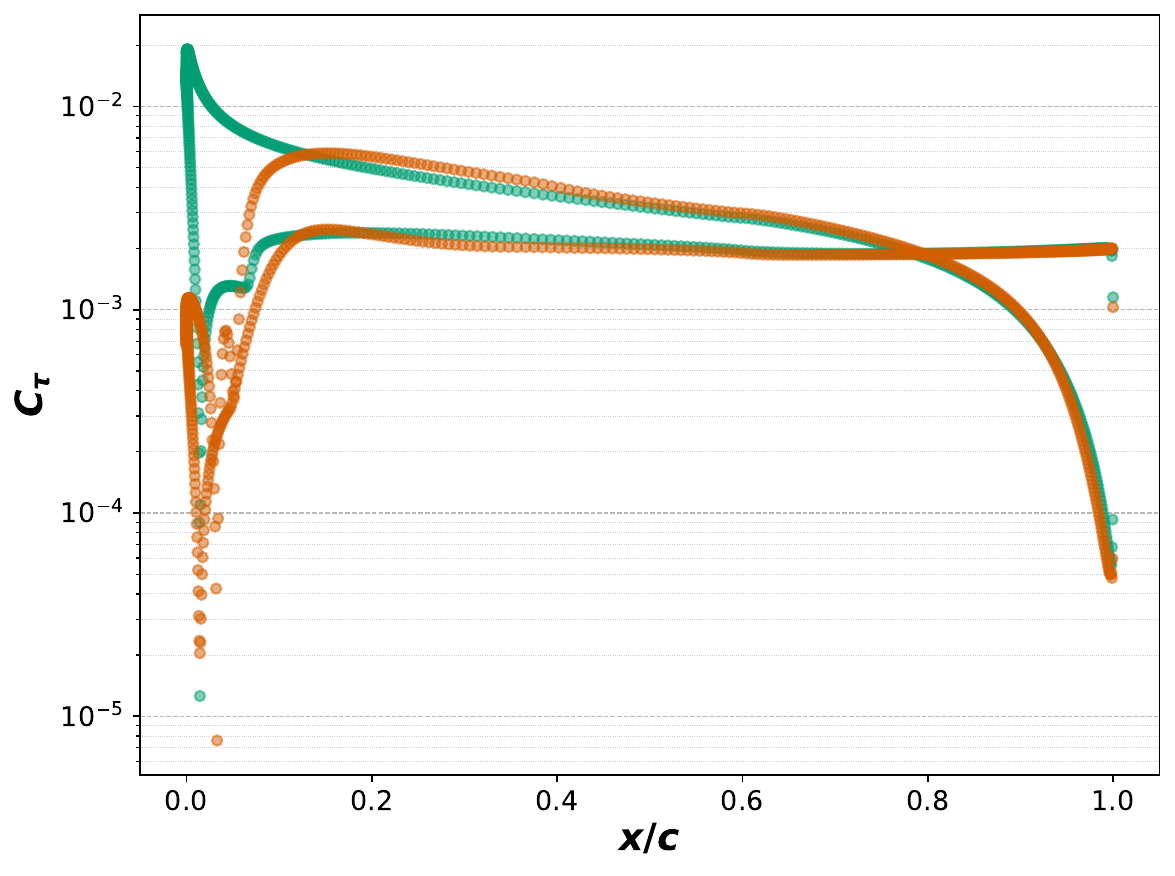}
        \caption{Skin friction coefficient.}
        \label{fig:af_sub3}
    \end{subfigure}

    \caption{Visual comparison of a low- and high-fidelity simulation of a NACA4 airfoil with parameters (\texttt{M=2.408, P=5.987, XX=11.876}), at an \ac{AoA} of \SI{7.57}{\degree} with an inlet velocity of \SI{81.645}{\metre\per\second}. 
    \cref{fig:af_sub1} (left) shows the $x$-velocity of the high-fidelity simulation, and (right) zooms on the corresponding boundary layer profile at mid-chord (\(0.5c\)), highlighting the difference in modeling resolution between low- and high-fidelity.
    Bottom plots display the evolution of the pressure coefficient $C_p$ (\cref{fig:af_sub2}) and skin friction coefficient $C_{\tau}$ (\cref{fig:af_sub3}), along the chord line $x/c$.}
    \label{fig:airfoil_plot_lf_hf}
\end{figure}

The numerical solution of PDEs depends on assumptions and choices made while designing the simulation pipeline, aiming to balance accuracy with computational feasibility.
In \ac{CFD} for external aerodynamics the goal is to solve the Navier-Stokes equations for the flow around rigid bodies.
Given this problem setting, the following design choices can be made when simulating the system:
\begin{enumerate}
    \item \textbf{Problem definition:} Define the high-level description of the system. For example, is the flow laminar or turbulent, compressible or incompressible, subsonic, transonic or supersonic. This also includes whether the problem should be solved in two or three dimensions and whether transient solutions are required or steady-state averages are sufficient. 
    \item \textbf{Physics modeling assumptions:} Choose the appropriate technique (e.g., \ac{LES}, \ac{RANS}, or hybrid methods). If turbulence is present, select a closure (e.g., one-equation models, two-equation models, etc.) and pick boundary layer treatment (fully resolved, wall functions).
    \item \textbf{Initial and boundary conditions:} Set inflow, outlet and surface boundary conditions, as well as initial conditions of the system.
    \item \textbf{Meshing:} Generate a mesh fine enough to support the modeling assumptions and simplifications defined in previous steps.
    \item \textbf{Solver settings:} Select discretization schemes, time-stepping methods, relaxation factors and convergence criteria.
\end{enumerate}

To study scaling, we select a dataset with the following criteria:
\begin{enumerate*}[label=(\roman*)]
    \item the problem setup should be realistic and not purely academic,
    \item the low- and high-fidelity datasets should differ in their physical modeling assumptions, not simply in mesh resolution, and
    \item the computational cost of the high-fidelity simulations should be noticeably larger than the low-fidelity simulations.
\end{enumerate*}

We identify aerodynamic airfoil simulations as an ideal testbed, since they are industrially relevant, well studied and allow for different fidelity levels based on physical modeling assumptions.
Due to the prohibitive cost of \ac{DNS} or \ac{LES} simulations for large dataset creation \citep{choimoin2021grid, yang2021grid}, we base our study on simulating \ac{RANS} equations.

In \ac{CFD} and especially external aerodynamics, the boundary layer, i.e. the thin region of fluid close to the solid's surface, is of utmost importance.
It is common to use a dimensionless wall distance $y^+$ (pronounced ``y-plus'') to describe the distance to the surface.
The region where $y^+<5$ corresponds to the viscous sublayer.
This layer is characterized by strong velocity gradients, and its accurate prediction is critical since key aerodynamic quantities, such as drag and lift, depend on these gradients.
To create two distinct fidelity levels, we vary the boundary layer treatment.
Our high-fidelity simulations \textit{fully resolve} this region by ensuring that the first computational mesh cell center has $y^+<1$, leading to accurate but costly predictions.
On the other hand, the low-fidelity setup uses a coarser mesh near the wall such that $y^+$ lies between $30$ and $300$.
This allows for a wall function approach, in which the region close to the wall is not resolved directly but \textit{modeled analytically} using the empirical law-of-the-wall derived from experimental data \citep{schlichting2016boundary}.
This modeling choice is widely used in RANS and wall-modeled LES (WMLES) applications.

We base our high-fidelity simulation setup on the AirfRANS dataset configuration \citep{bonnet2022airfrans}.
AirfRANS models airfoils from the NASA's 4- and 5-digit series \citep{cummings2015applied} in an incompressible regime (Mach number $<0.3$), covering Reynolds numbers from $2\times10^6$ to $6\times10^6$, and \acp{AoA} between $-5^\circ$ and $15^\circ$.
We run all simulations in OpenFOAM \citep{weller1998openfoam}, using the \texttt{simpleFOAM} solver, with the $k$--$\omega$ SST turbulence model \citep{menter2003sst} as equation closure (a standard approach in airfoil aerodynamics).
To highlight the important aspects of the resulting datasets, \cref{tab:simulation_specs} summarizes the differences between high- and low-fidelity.
All numerical simulations were run on an AMD EPYC9655P 96-Core CPU (192 threads, 4.5GHz and 2.2 TiB of RAM). We use OpenFOAMv2506 and Open MPI 4.1.1 for parallel execution.

\begin{table}[t]
\centering
\renewcommand{\arraystretch}{1.3}
\setlength{\tabcolsep}{6pt} 
\caption{Low- and high-fidelity modeling assumptions and resulting dataset characteristics.\label{tab:simulation_specs}\vspace{6px}}
\begin{tabular}{@{}lccc ccc@{}}
\toprule
\multicolumn{4}{c}{\textbf{Modeling Assumptions}} & \multicolumn{3}{c}{\textbf{Dataset Specifications}} \\
\cmidrule(r){1-4} \cmidrule(l){5-7}
Fidelity & \makecell{Viscous \\ sublayer} & \makecell{First cell \\ height ($\mu$m)} & \makecell{First cell \\ center ($y^+$)} 
& \makecell{Avg sim time \\ (core hrs)} & \makecell{Avg number \\ of nodes} & \makecell{Total size \\ (GB)} \\
\cmidrule(r){1-4} \cmidrule(l){5-7}
High & Resolving & 2     & $<1$        & 13.4   & 180K & 18 \\
Low  & Modeling  & 1,200 & 30--300    & 4.8    & 96K  & 7.8 \\
\bottomrule
\end{tabular}
\end{table}

The final datasets consist of 611 matched pairs of low- and high-fidelity simulations (491 train/val, 120 test).
The difference in dataset size compared to the original AirfRANS dataset is caused by the simplifications needed for the low-fidelity simulations, which can sometimes lead to poor convergence. 
\cref{fig:airfoil_plot_lf_hf} illustrates the difference between the two fidelities in terms of boundary layer profile and distribution of key physical quantities for a chosen dataset sample.

\section{Experiments}
\label{sec:experiments}
Our experiments are designed to investigate \textit{compute-optimal} model training in the setting where the available budget $D_b$ for data generation is the primary constraint.
Model size is fixed across all experiments, and our analysis focuses on the optimal composition of low- and high-fidelity data. For fixed $D_b$s (in core hours) we vary the ratio of low- to high-fidelity samples $D_c$ in the training set.
This is done by first estimating the number of datapoints based on the average cost and the desired fidelity distribution, then sampling random low-/high-fidelity simulations until $D_b$ is matched, and finally applying an optional greedy repair step to ensure the final selection satisfies the budget constraint.

The task at hand is to predict the solution of the \ac{RANS} simulation given the initial conditions and mesh node positions.
For all nodes, we predict six quantities: velocity $v$ in $x$- and $y$-direction, pressure $p$, and wall shear stress $\tau$ in  $x$- and $y$-direction.
We use Transolver \citep{wu2024transolver}, a SOTA transformer-based neural operator, with $\sim$4M parameters. Detailed model and training hyperparameters are provided in \cref{app:training_details}.

\begin{figure}[htbp]
    \centering

    \begin{subfigure}[b]{0.45\textwidth}
        \centering
        \includegraphics[width=\linewidth]{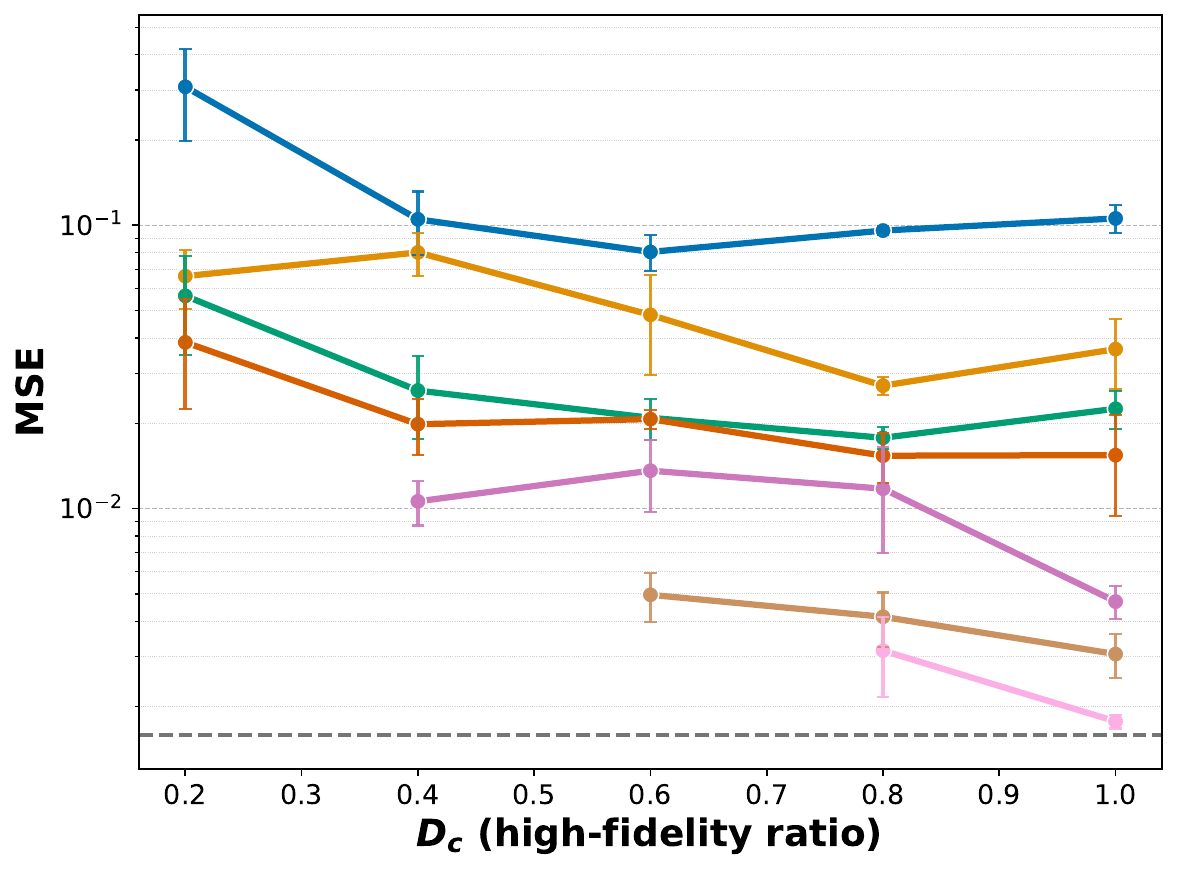}
        \caption{Volumetric pressure}
        \label{fig:mse_vol_p}
    \end{subfigure}
    \hfill
    \begin{subfigure}[b]{0.45\textwidth}
        \centering
        \includegraphics[width=\linewidth]{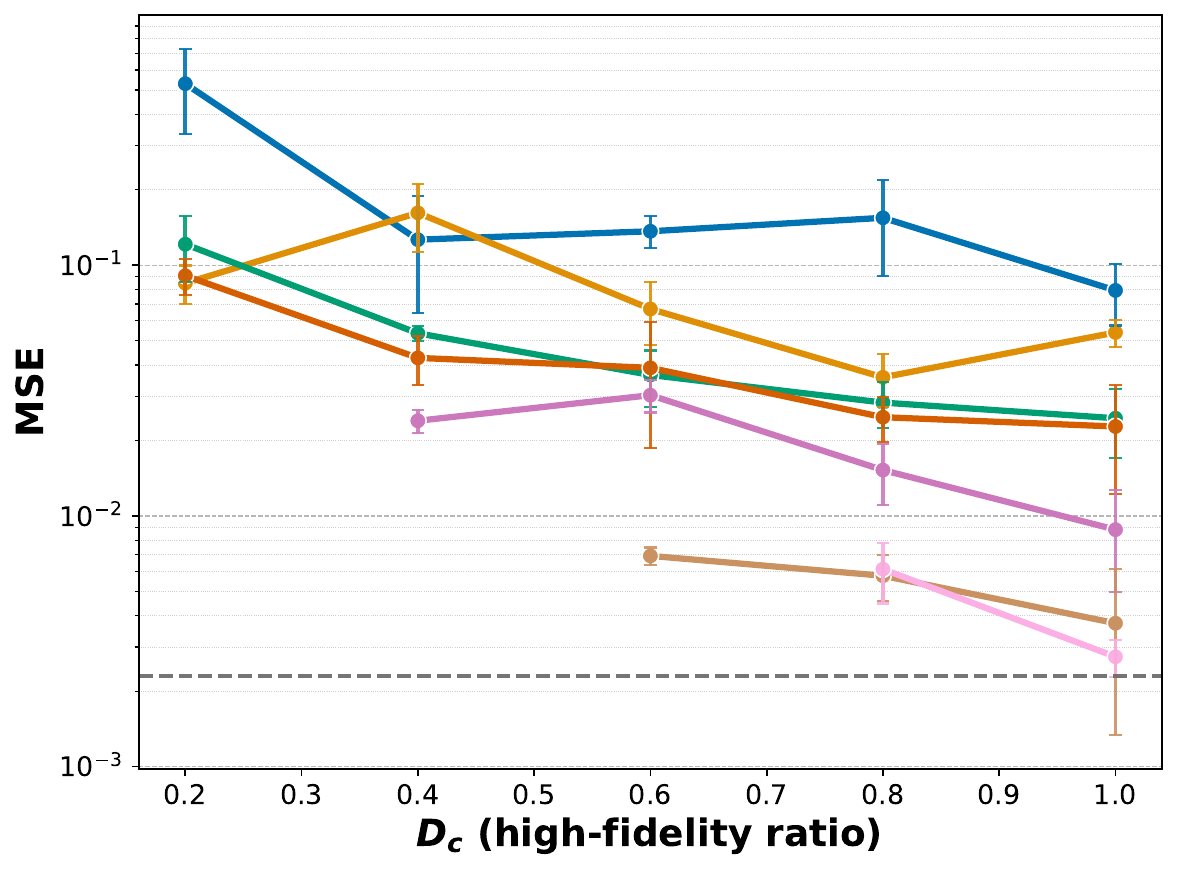}
        \caption{Surface pressure}
        \label{fig:mse_surf_p}
    \end{subfigure}


    \vspace{-1.2em}

    \begin{subfigure}[b]{0.3\textwidth}
        \centering
        \includegraphics[width=\linewidth]{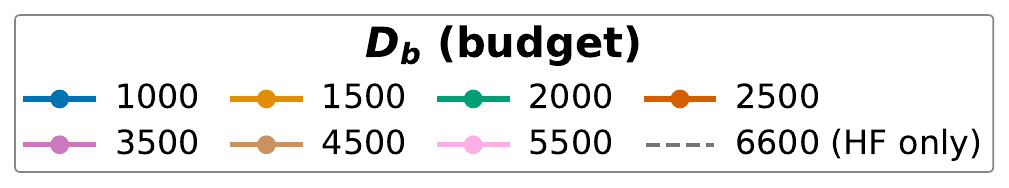}
    \end{subfigure}

    \vspace{0.3em}

    \begin{subfigure}[b]{0.45\textwidth}
        \centering
        \includegraphics[width=\linewidth]{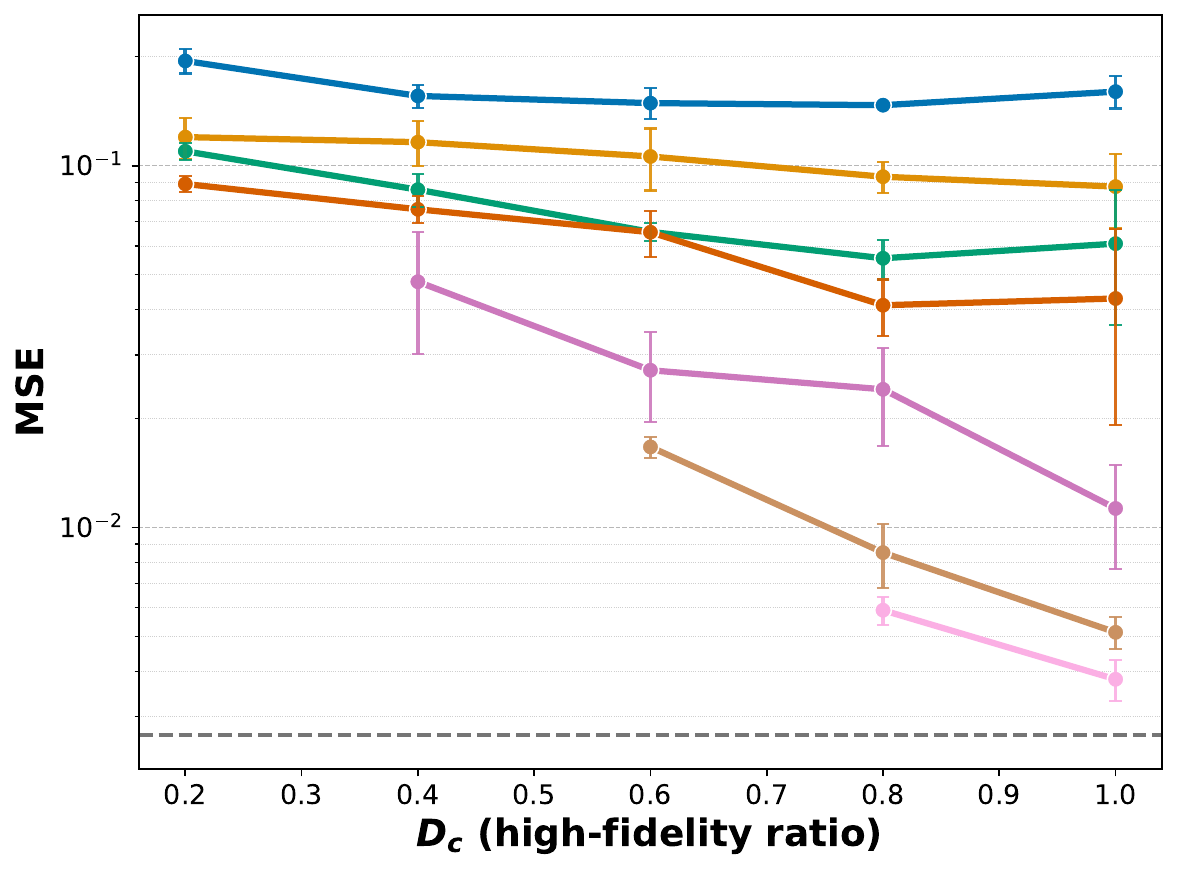}
        \caption{Volumetric $x$-velocity}
        \label{fig:mse_vol_vel_x}
    \end{subfigure}
    \hfill
    \begin{subfigure}[b]{0.45\textwidth}
        \centering
        \includegraphics[width=\linewidth]{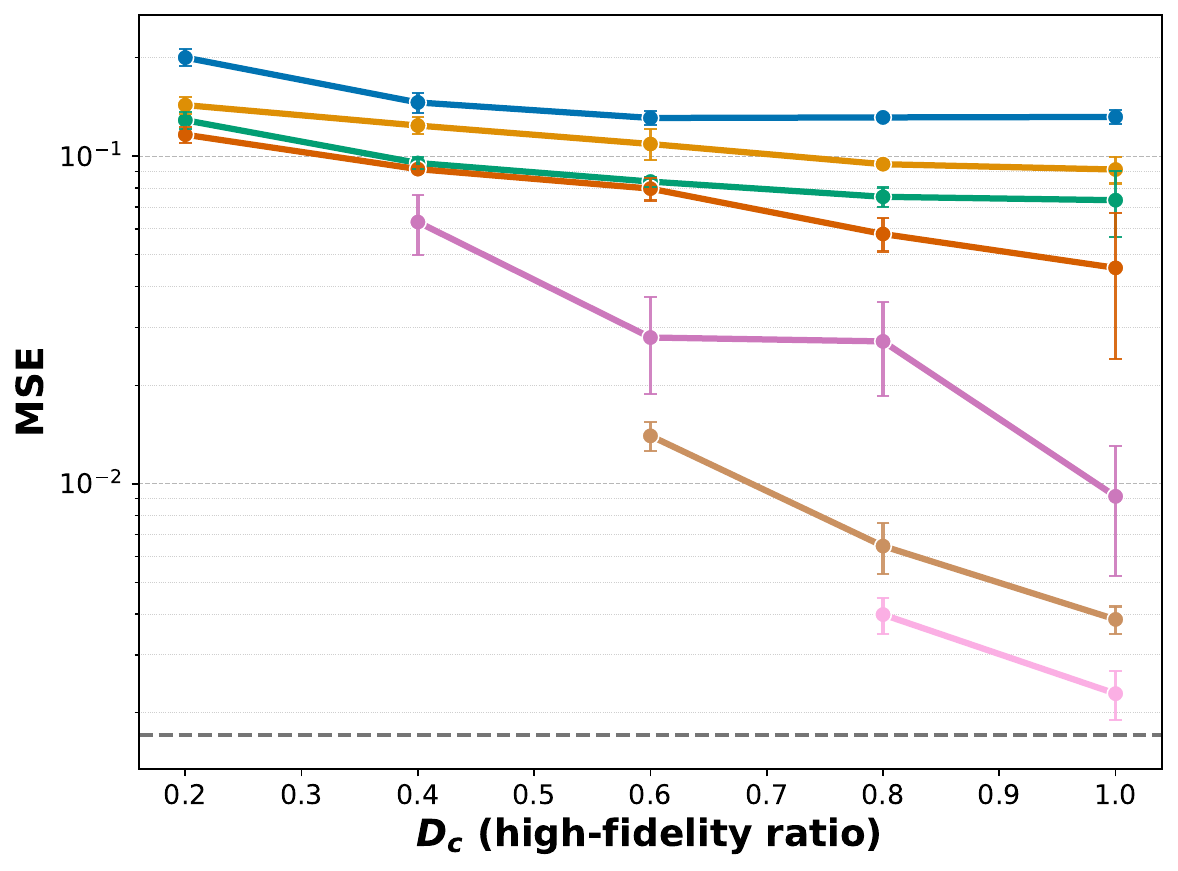}
        \caption{Volumetric $y$-velocity}
        \label{fig:mse_vol_vel_y}
    \end{subfigure}
    \caption{Scaling behavior for fields with positive transfer. We show the \ac{MSE} of normalized fields averaged over four seeds with error bars indicating standard deviation. Lines of different colors show different training budgets in compute hours $D_b$, at growing percentage of high-fidelity composition $D_c$. The dashed line indicates model performance when trained on the full high-fidelity dataset.}
    \label{fig:scaling_plots_pos_transfer}
\end{figure}

\begin{figure}[htbp]
    \centering

    \begin{subfigure}[b]{0.3\textwidth}
        \centering
        \includegraphics[width=\linewidth]{figs/legend_budgets.pdf}
    \end{subfigure}

    \vspace{0.3em}

    \begin{subfigure}[b]{0.45\textwidth}
        \centering
        \includegraphics[width=\linewidth]{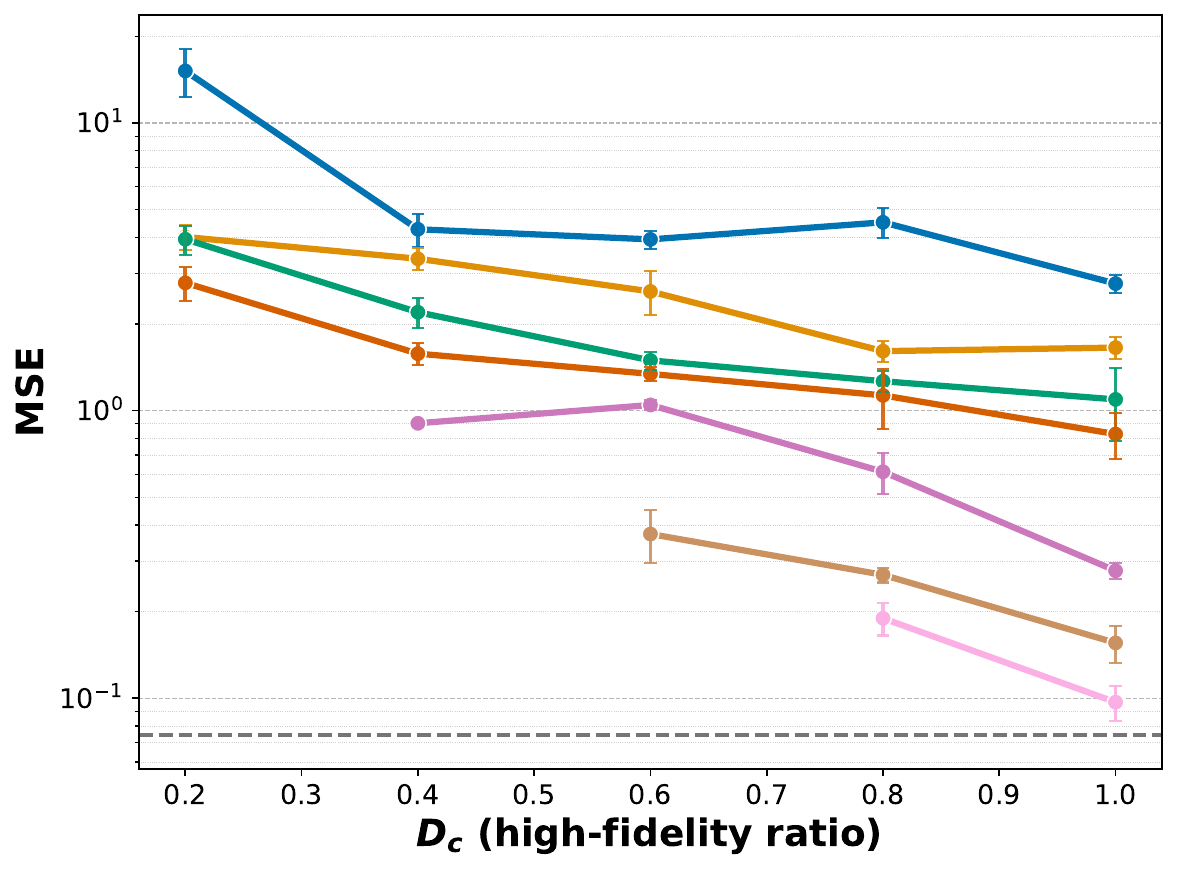}
        \caption{Wall shear stress ($x$-component)}
        \label{fig:mse_surf_wss_x}
    \end{subfigure}
    \hfill
    \begin{subfigure}[b]{0.45\textwidth}
        \centering
        \includegraphics[width=\linewidth]{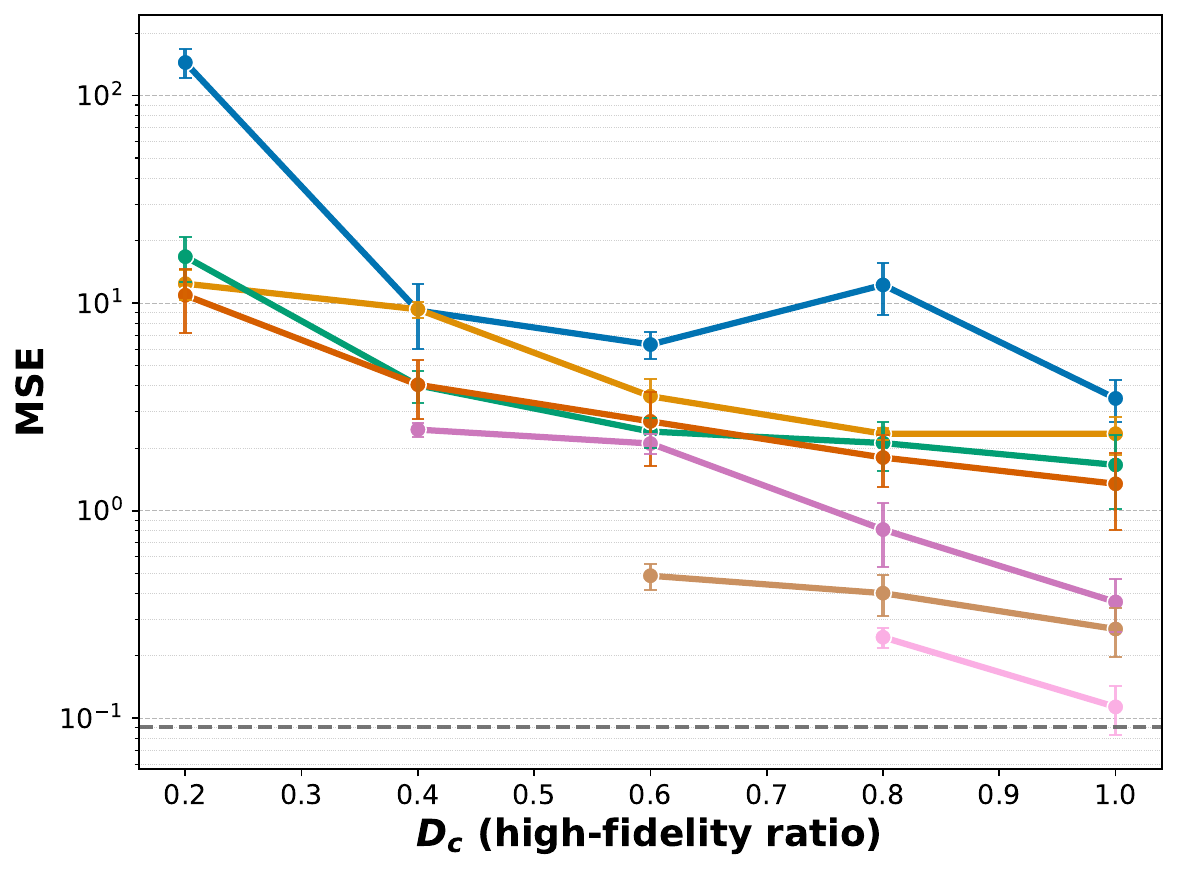}
        \caption{Wall shear stress ($y$-component)}
        \label{fig:mse_surf_wss_y}
    \end{subfigure}

    \caption{Scaling behavior for fields without positive transfer. We show the \ac{MSE} of normalized fields averaged over four seeds with error bars indicating standard deviation. Lines of different colors show different training budgets in compute hours $D_b$, at growing percentage of high-fidelity composition $D_c$. The dashed line indicates model performance when trained on the full high-fidelity dataset.}
    \label{fig:scaling_plots_neg_transfer}
\end{figure}

The same architecture is trained on different datasets, generated at different budgets $D_b$ and with varying ratios of low- and high-fidelity simulations $D_c$.
Performance is evaluated on 120 unseen high-fidelity samples.
\cref{fig:scaling_plots_pos_transfer} and \cref{fig:scaling_plots_neg_transfer} show the results of how model performance behaves with increasing dataset generation budgets and varying dataset compositions.
We discuss our main findings below.

\paragraph{Compute Budget Scaling Law.}
Across all budgets and dataset compositions, we observe a clear trend that model error decreases with an increasing compute budget for dataset generation (\cref{fig:scaling_plots_pos_transfer,fig:scaling_plots_neg_transfer}).
This confirms that the used budget for training data simulations directly links to surrogate accuracy, analogous to scaling laws observed in model, data and compute size in other domains \citep{kaplan2020scaling,hoffman2022chinchilla}.

\paragraph{Knowledge Transfer from Low- to High-Fidelity.}
For lower compute budgets, certain fields show signs of positive transfer from low- to high-fidelity samples.
This behavior is visible for the pressure field in the volume (\cref{fig:mse_vol_p}) and on the surface (\cref{fig:mse_surf_p}) as well as the volumetric velocity field (\cref{fig:mse_vol_vel_x,fig:mse_vol_vel_y}).
When the available dataset generation budget is limited, allocating all resources to high-fidelity samples does not lead to optimal test performance.
Instead, models trained on a mixture of low- and high-fidelity data achieve better accuracy.
This suggests that, \textit{under tight compute constraints, the broader coverage of the data manifold offered by many low-fidelity samples outweighs the higher accuracy of a few high-fidelity ones.}
In general, the smaller the available budget, the more the optimal dataset composition shifts towards allocating more budget to lower fidelity samples.
Above certain budgets, model performance continuously improves with more budget allocated towards high-fidelity samples, showing that beyond a certain budget threshold, model accuracy becomes primarily limited by the fidelity of the data rather than its quantity.

Contrary to these trends, \textit{we cannot observe any positive transfer from low- to high-fidelity samples for the wall shear stress.}
\cref{fig:mse_surf_wss_x,fig:mse_surf_wss_y} show that for these quantities model performance consistently improves across all budgets when more dataset generation budget is allocated towards high-fidelity samples.

\begin{wraptable}{r}{0.42\textwidth}
\vspace{-\baselineskip} 
\centering
\renewcommand{\arraystretch}{1.1}
\setlength{\tabcolsep}{6pt}
\caption{\acs{nMAE} ($\downarrow$) per field between low- and high-fidelity simulations.}
\label{tab:diff_hf_lf}
\begin{tabular}{lcc}
\toprule
\textbf{Field} & \textbf{Surface} & \textbf{Volume} \\
\midrule
($x$-)Velocity & --    & 0.118 \\
($y$-)Velocity & --    & 0.303 \\
Pressure       & \textbf{0.043} & \textbf{0.040} \\
($x$-)WSS      & 0.405 & -- \\
($y$-)WSS      & 0.796 & -- \\
\bottomrule
\end{tabular}
\end{wraptable}

\paragraph{Physical Explanation.}
We hypothesize that the observed results can be linked to the discrepancies between the low- and high-fidelity simulations.
The primary distinction lies in the treatment of the viscous sublayer at the airfoil surface: the low-fidelity setup models this region using relatively coarse meshing, whereas the high-fidelity simulation fully resolves it with a fine mesh.
As a result, velocity and pressure fields remain largely consistent across fidelities, while the wall shear stress, which is highly sensitive to the boundary layer resolution, shows substantial deviations.

This explains why no positive transfer can be observed for this quantity even at small dataset generation budgets: the difference between the two simulations is simply too large.
\cref{tab:diff_hf_lf} quantifies these discrepancies per field by reporting the \ac{nMAE} (\cref{app:nmae}) of low-fidelity fields interpolated onto the corresponding high-fidelity mesh relative to their high-fidelity counterparts.
It clearly shows a difference in \ac{nMAE} for the pressure field compared to the two components of wall shear stress, aligning with the different multi-fidelity scaling behaviors shown in \cref{fig:mse_surf_p,fig:mse_vol_p} compared to \cref{fig:mse_surf_wss_x,fig:mse_surf_wss_y}.
This also aligns with the visual comparison of the difference in pressure coefficients $C_p$ and skin friction coefficients $C_{\tau}$ along the chord (see \cref{fig:af_sub2,fig:af_sub3}).

\section{Conclusion and Future Work}
\label{sec:future_work}
Our work serves as an initial step towards understanding \textit{scaling laws for neural surrogates trained on multi-fidelity data}, highlighting both the potential of optimal dataset budget allocation and the limitations arising when the fidelity gap between simulations becomes too large.
Given our findings, we identify several promising directions.

\paragraph{Different simulation methods.}
Our experiments are currently limited to \ac{RANS} simulations where fidelity is varied via boundary layer treatment.
Exploring additional simulation methods as fidelities, such as time-averaged LES or hybrid \ac{RANS}-\ac{LES} approaches, could provide deeper insights into realistic multi-fidelity dataset design, albeit at increased computational cost.

\paragraph{Continuous fidelities.}
Our results support the development of continuous fidelity formulations rather than discrete fidelity levels.
This is both more realistic, since every simulation inherently allows continuous mesh scaling, and potentially more effective, as it can mitigate situations where fidelity levels are too far apart for meaningful knowledge transfer.

\paragraph{Generalization of the framework.}
Extending our multi-fidelity scaling analysis to other scientific domains, such as thermomechanics, electromagnetics, or molecular dynamics, could reveal whether the identified scaling behaviors generalize across different physical systems.
Additionally, while our study focuses on dataset generation cost and composition, future work should also explore scaling the remaining axes, namely model size and training compute in order to eventually establish a more complete formulation of multi-fidelity scientific scaling laws.

\section*{Acknowledgments}
The authors thank Fabian Paischer for the insightful conversations and the constructive feedback, and L\'eo Cotteleer for the discussions about the numerical simulations.

The ELLIS Unit Linz, the LIT AI Lab, the Institute for Machine Learning, are supported by the Federal State Upper Austria. We thank the projects FWF AIRI FG 9-
N (10.55776/FG9), AI4GreenHeatingGrids (FFG- 899943), Stars4Waters (HORIZON-CL6-2021-
CLIMATE-01-01), FWF Bilateral Artificial Intelligence (10.55776/COE12). We acknowledge EuroHPC Joint Undertaking for awarding us access to Leonardo at CINECA, Italy, and MareNostrum5 at BSC, Spain.

\newpage
\bibliographystyle{neurips_2025}
\bibliography{references}

\begin{thebibliography}{47}
\providecommand{\natexlab}[1]{#1}
\providecommand{\url}[1]{\texttt{#1}}
\expandafter\ifx\csname urlstyle\endcsname\relax
  \providecommand{\doi}[1]{doi: #1}\else
  \providecommand{\doi}{doi: \begingroup \urlstyle{rm}\Url}\fi

\bibitem[Abramson et~al.(2024)Abramson, Adler, Dunger, Evans, Green, Pritzel, Ronneberger, Willmore, Ballard, Bambrick, et~al.]{abramson2024alphafold3}
Josh Abramson, Jonas Adler, Jack Dunger, Richard Evans, Tim Green, Alexander Pritzel, Olaf Ronneberger, Lindsay Willmore, Andrew~J Ballard, Joshua Bambrick, et~al.
\newblock Accurate structure prediction of biomolecular interactions with alphafold 3.
\newblock \emph{Nature}, pp.\  1--3, 2024.

\bibitem[AI(2025)]{mistral2025large2}
Mistral AI.
\newblock Mistral large 2 release announcement.
\newblock \url{https://mistral.ai/news/mistral-large-2407}, 2025.

\bibitem[Alkin et~al.(2024)Alkin, F{\"{u}}rst, Schmid, Gruber, Holzleitner, and Brandstetter]{alkin2024upt}
Benedikt Alkin, Andreas F{\"{u}}rst, Simon Schmid, Lukas Gruber, Markus Holzleitner, and Johannes Brandstetter.
\newblock Universal physics transformers: {A} framework for efficiently scaling neural operators.
\newblock In Amir Globersons, Lester Mackey, Danielle Belgrave, Angela Fan, Ulrich Paquet, Jakub~M. Tomczak, and Cheng Zhang (eds.), \emph{Advances in Neural Information Processing Systems 38: Annual Conference on Neural Information Processing Systems 2024, NeurIPS 2024, Vancouver, BC, Canada, December 10 - 15, 2024}, 2024.
\newblock URL \url{http://papers.nips.cc/paper\_files/paper/2024/hash/2cd36d327f33d47b372d4711edd08de0-Abstract-Conference.html}.

\bibitem[Ashton et~al.(2024)Ashton, Mockett, Fuchs, Fliessbach, Hetmann, Knacke, Schonwald, Skaperdas, Fotiadis, Walle, Hupertz, and Maddix]{ashton2024drivearml}
Neil Ashton, Charles Mockett, Marian Fuchs, Louis Fliessbach, Hendrik Hetmann, Thilo Knacke, Norbert Schonwald, Vangelis Skaperdas, Grigoris Fotiadis, Astrid Walle, Burkhard Hupertz, and Danielle~C. Maddix.
\newblock Drivaerml: High-fidelity computational fluid dynamics dataset for road-car external aerodynamics.
\newblock \emph{CoRR}, abs/2408.11969, 2024.
\newblock \doi{10.48550/ARXIV.2408.11969}.
\newblock URL \url{https://doi.org/10.48550/arXiv.2408.11969}.

\bibitem[Bartolucci et~al.(2023)Bartolucci, de~Bézenac, Raonić, Molinaro, Mishra, and Alaifari]{bartolucci2023representationequivalentneuraloperators}
Francesca Bartolucci, Emmanuel de~Bézenac, Bogdan Raonić, Roberto Molinaro, Siddhartha Mishra, and Rima Alaifari.
\newblock Representation equivalent neural operators: a framework for alias-free operator learning, 2023.
\newblock URL \url{https://arxiv.org/abs/2305.19913}.

\bibitem[Berman et~al.(2000)Berman, Westbrook, Feng, Gilliland, Bhat, Weissig, Shindyalov, and Bourne]{berman2000pdb}
Helen~M. Berman, John Westbrook, Zukang Feng, Gary Gilliland, Talapady~N. Bhat, Helge Weissig, Ilya~N. Shindyalov, and Philip~E. Bourne.
\newblock The protein data bank.
\newblock \emph{Nucleic Acids Research}, 28\penalty0 (1):\penalty0 235--242, 2000.
\newblock \doi{10.1093/nar/28.1.235}.

\bibitem[Beyer et~al.(2023)Beyer, Izmailov, Kolesnikov, Caron, Kornblith, Zhai, Minderer, Tschannen, Alabdulmohsin, and Pavetic]{beyer2023flexivitmodelpatchsizes}
Lucas Beyer, Pavel Izmailov, Alexander Kolesnikov, Mathilde Caron, Simon Kornblith, Xiaohua Zhai, Matthias Minderer, Michael Tschannen, Ibrahim Alabdulmohsin, and Filip Pavetic.
\newblock Flexivit: One model for all patch sizes, 2023.
\newblock URL \url{https://arxiv.org/abs/2212.08013}.

\bibitem[Bodnar et~al.(2025)Bodnar, Bruinsma, Lucic, Stanley, Allen, Brandstetter, Garvan, Riechert, Weyn, Dong, Gupta, Thambiratnam, Archibald, Wu, Heider, Welling, Turner, and Perdikaris]{bodnar2024aurora}
Cristian Bodnar, Wessel~P. Bruinsma, Ana Lucic, Megan Stanley, Anna Allen, Johannes Brandstetter, Patrick Garvan, Maik Riechert, Jonathan~A. Weyn, Haiyu Dong, Jayesh~K. Gupta, Kit Thambiratnam, Alexander~T. Archibald, Chun{-}Chieh Wu, Elizabeth Heider, Max Welling, Richard~E. Turner, and Paris Perdikaris.
\newblock A foundation model for the earth system.
\newblock \emph{Nat.}, 641\penalty0 (8065):\penalty0 1180--1187, 2025.
\newblock \doi{10.1038/S41586-025-09005-Y}.
\newblock URL \url{https://doi.org/10.1038/s41586-025-09005-y}.

\bibitem[Bonnet et~al.(2022)Bonnet, Mazari, Cinnella, and Gallinari]{bonnet2022airfrans}
Florent Bonnet, Jocelyn~Ahmed Mazari, Paola Cinnella, and Patrick Gallinari.
\newblock Airfrans: High fidelity computational fluid dynamics dataset for approximating reynolds-averaged navier-stokes solutions.
\newblock In Sanmi Koyejo, S.~Mohamed, A.~Agarwal, Danielle Belgrave, K.~Cho, and A.~Oh (eds.), \emph{Advances in Neural Information Processing Systems 35: Annual Conference on Neural Information Processing Systems 2022, NeurIPS 2022, New Orleans, LA, USA, November 28 - December 9, 2022}, 2022.
\newblock URL \url{http://papers.nips.cc/paper\_files/paper/2022/hash/94ab7b23a345f93333eac8748a66c763-Abstract-Datasets\_and\_Benchmarks.html}.

\bibitem[Choi \& Moin(2012)Choi and Moin]{choimoin2021grid}
Haecheon Choi and Parviz Moin.
\newblock Grid-point requirements for large eddy simulation: Chapman’s estimates revisited.
\newblock \emph{Physics of Fluids}, 24, 01 2012.
\newblock \doi{10.1063/1.3676783}.

\bibitem[Cummings et~al.(2015)Cummings, Mason, Morton, and McDaniel]{cummings2015applied}
Russell~M. Cummings, William~H. Mason, Scott~A. Morton, and David~R. McDaniel.
\newblock Applied computational aerodynamics: A modern engineering approach.
\newblock In \emph{Cambridge Aerospace Series}, pp.\  731--765. Cambridge University Press, 2015.
\newblock \doi{10.1017/CBO9781107284166}.

\bibitem[De et~al.(2020)De, Britton, Reynolds, Skinner, Jansen, and Doostan]{de2020transfer_learning_mf}
Subhayan De, Jolene Britton, Matthew~J. Reynolds, Ryan Skinner, Kenneth~E. Jansen, and Alireza Doostan.
\newblock On transfer learning of neural networks using bi-fidelity data for uncertainty propagation.
\newblock \emph{CoRR}, abs/2002.04495, 2020.
\newblock URL \url{https://arxiv.org/abs/2002.04495}.

\bibitem[Dong et~al.(2015)Dong, Loy, He, and Tang]{dong2015imagesuperresolutionusingdeep}
Chao Dong, Chen~Change Loy, Kaiming He, and Xiaoou Tang.
\newblock Image super-resolution using deep convolutional networks, 2015.
\newblock URL \url{https://arxiv.org/abs/1501.00092}.

\bibitem[Elrefaie et~al.(2024)Elrefaie, Morar, Dai, and Ahmed]{elrefaie2024drivearnetpp}
Mohamed Elrefaie, Florin Morar, Angela Dai, and Faez Ahmed.
\newblock Drivaernet++: A large-scale multimodal car dataset with computational fluid dynamics simulations and deep learning benchmarks.
\newblock In A.~Globerson, L.~Mackey, D.~Belgrave, A.~Fan, U.~Paquet, J.~Tomczak, and C.~Zhang (eds.), \emph{Advances in Neural Information Processing Systems}, volume~37, pp.\  499--536. Curran Associates, Inc., 2024.
\newblock URL \url{https://proceedings.neurips.cc/paper_files/paper/2024/file/013cf29a9e68e4411d0593040a8a1eb3-Paper-Datasets_and_Benchmarks_Track.pdf}.

\bibitem[Evans(2010)]{evans2010pde}
Lawrence~C. Evans.
\newblock \emph{Partial Differential Equations}, volume~19 of \emph{Graduate Studies in Mathematics}.
\newblock American Mathematical Society, 2nd edition, 2010.

\bibitem[Herde et~al.(2024)Herde, Raonic, Rohner, K{\"{a}}ppeli, Molinaro, de~B{\'{e}}zenac, and Mishra]{herde2024poseidon}
Maximilian Herde, Bogdan Raonic, Tobias Rohner, Roger K{\"{a}}ppeli, Roberto Molinaro, Emmanuel de~B{\'{e}}zenac, and Siddhartha Mishra.
\newblock Poseidon: Efficient foundation models for pdes.
\newblock In Amir Globersons, Lester Mackey, Danielle Belgrave, Angela Fan, Ulrich Paquet, Jakub~M. Tomczak, and Cheng Zhang (eds.), \emph{Advances in Neural Information Processing Systems 38: Annual Conference on Neural Information Processing Systems 2024, NeurIPS 2024, Vancouver, BC, Canada, December 10 - 15, 2024}, 2024.
\newblock URL \url{http://papers.nips.cc/paper\_files/paper/2024/hash/84e1b1ec17bb11c57234e96433022a9a-Abstract-Conference.html}.

\bibitem[Hersbach et~al.(2023)Hersbach, Bell, Berrisford, Biavati, Horányi, Muñoz~Sabater, Nicolas, Peubey, Radu, Rozum, Schepers, Simmons, Soci, Dee, and Thépaut]{hersbach2023era5}
Hans Hersbach, Bill Bell, Paul Berrisford, Guido Biavati, András Horányi, Joaquín Muñoz~Sabater, Julien Nicolas, Carole Peubey, Raluca Radu, Iryna Rozum, Dinand Schepers, Adrian Simmons, Cătălin Soci, Dick Dee, and Jean-Noël Thépaut.
\newblock {ERA5 hourly data on single levels from 1940 to present}, 2023.
\newblock URL \url{https://doi.org/10.24381/cds.adbb2d47}.
\newblock Accessed on DD-MMM-YYYY.

\bibitem[Hoffmann et~al.(2022)Hoffmann, Borgeaud, Mensch, Buchatskaya, Cai, Rutherford, de~Las~Casas, Hendricks, Welbl, Clark, Hennigan, Noland, Millican, van~den Driessche, Damoc, Guy, Osindero, Simonyan, Elsen, Rae, Vinyals, and Sifre]{hoffman2022chinchilla}
Jordan Hoffmann, Sebastian Borgeaud, Arthur Mensch, Elena Buchatskaya, Trevor Cai, Eliza Rutherford, Diego de~Las~Casas, Lisa~Anne Hendricks, Johannes Welbl, Aidan Clark, Tom Hennigan, Eric Noland, Katie Millican, George van~den Driessche, Bogdan Damoc, Aurelia Guy, Simon Osindero, Karen Simonyan, Erich Elsen, Jack~W. Rae, Oriol Vinyals, and Laurent Sifre.
\newblock Training compute-optimal large language models.
\newblock \emph{CoRR}, abs/2203.15556, 2022.
\newblock \doi{10.48550/ARXIV.2203.15556}.
\newblock URL \url{https://doi.org/10.48550/arXiv.2203.15556}.

\bibitem[Jumper et~al.(2021)Jumper, Evans, Pritzel, Green, Figurnov, Ronneberger, Tunyasuvunakool, Bates, {\v{Z}}{\'\i}dek, Potapenko, et~al.]{jumper2021alphafold}
John Jumper, Richard Evans, Alexander Pritzel, Tim Green, Michael Figurnov, Olaf Ronneberger, Kathryn Tunyasuvunakool, Russ Bates, Augustin {\v{Z}}{\'\i}dek, Anna Potapenko, et~al.
\newblock Highly accurate protein structure prediction with alphafold.
\newblock \emph{nature}, 596\penalty0 (7873):\penalty0 583--589, 2021.

\bibitem[Kaplan et~al.(2020)Kaplan, McCandlish, Henighan, Brown, Chess, Child, Gray, Radford, Wu, and Amodei]{kaplan2020scaling}
Jared Kaplan, Sam McCandlish, Tom Henighan, Tom~B. Brown, Benjamin Chess, Rewon Child, Scott Gray, Alec Radford, Jeffrey Wu, and Dario Amodei.
\newblock Scaling laws for neural language models.
\newblock \emph{CoRR}, abs/2001.08361, 2020.
\newblock URL \url{https://arxiv.org/abs/2001.08361}.

\bibitem[Keisler(2022)]{keisler2022forecasting_global_weather}
Ryan Keisler.
\newblock Forecasting global weather with graph neural networks, 2022.
\newblock URL \url{https://arxiv.org/abs/2202.07575}.

\bibitem[Kovachki et~al.(2021)Kovachki, Li, Liu, Azizzadenesheli, Bhattacharya, Stuart, and Anandkumar]{kovachki2021neural}
Nikola~B. Kovachki, Zongyi Li, Burigede Liu, Kamyar Azizzadenesheli, Kaushik Bhattacharya, Andrew~M. Stuart, and Anima Anandkumar.
\newblock Neural operator: Learning maps between function spaces.
\newblock \emph{CoRR}, abs/2108.08481, 2021.

\bibitem[Lopes et~al.(2024)Lopes, Puga, Teixeira, Lima, Grilo, Dueñas-Pamplona, and Ferrera]{Lopes2024lesvrrans}
D.~Lopes, H.~Puga, J.~Teixeira, R.~Lima, J.~Grilo, J.~Dueñas-Pamplona, and C.~Ferrera.
\newblock Comparison of rans and les turbulent flow models in a real stenosis.
\newblock \emph{International Journal of Heat and Fluid Flow}, 107:\penalty0 109340, 2024.
\newblock ISSN 0142-727X.
\newblock \doi{https://doi.org/10.1016/j.ijheatfluidflow.2024.109340}.
\newblock URL \url{https://www.sciencedirect.com/science/article/pii/S0142727X24000651}.

\bibitem[Loshchilov \& Hutter(2019)Loshchilov and Hutter]{loshchilov2019adamw}
Ilya Loshchilov and Frank Hutter.
\newblock Decoupled weight decay regularization.
\newblock In \emph{7th International Conference on Learning Representations, {ICLR} 2019, New Orleans, LA, USA, May 6-9, 2019}. OpenReview.net, 2019.
\newblock URL \url{https://openreview.net/forum?id=Bkg6RiCqY7}.

\bibitem[Lu et~al.(2022)Lu, Pestourie, Johnson, and Romano]{lu2022multifidelityneuraloperators}
Lu~Lu, Rapha{\"{e}}l Pestourie, Steven~G. Johnson, and Giuseppe Romano.
\newblock Multifidelity deep neural operators for efficient learning of partial differential equations with application to fast inverse design of nanoscale heat transport.
\newblock \emph{CoRR}, abs/2204.06684, 2022.
\newblock \doi{10.48550/ARXIV.2204.06684}.
\newblock URL \url{https://doi.org/10.48550/arXiv.2204.06684}.

\bibitem[Menter et~al.(2003)Menter, Kuntz, and Langtry]{menter2003sst}
Florian Menter, M.~Kuntz, and RB~Langtry.
\newblock Ten years of industrial experience with the sst turbulence model.
\newblock \emph{Heat and Mass Transfer}, 4, 01 2003.

\bibitem[Merchant et~al.(2023)Merchant, Batzner, Schoenholz, Aykol, Cheon, and Cubuk]{merchant2023scaling_dl_for_materials}
Amil Merchant, Simon~L. Batzner, Samuel~S. Schoenholz, Muratahan Aykol, Gowoon Cheon, and Ekin~Dogus Cubuk.
\newblock Scaling deep learning for materials discovery.
\newblock \emph{Nat.}, 624\penalty0 (7990):\penalty0 80--85, 2023.
\newblock \doi{10.1038/S41586-023-06735-9}.

\bibitem[Moser et~al.(2025)Moser, Shanbhag, Raue, Frolov, Palacio, and Dengel]{Moser_2025}
Brian~B. Moser, Arundhati~S. Shanbhag, Federico Raue, Stanislav Frolov, Sebastian Palacio, and Andreas Dengel.
\newblock Diffusion models, image super-resolution, and everything: A survey.
\newblock \emph{IEEE Transactions on Neural Networks and Learning Systems}, 36\penalty0 (7):\penalty0 11793–11813, July 2025.
\newblock ISSN 2162-2388.
\newblock \doi{10.1109/tnnls.2024.3476671}.
\newblock URL \url{http://dx.doi.org/10.1109/TNNLS.2024.3476671}.

\bibitem[Nguyen et~al.(2023)Nguyen, Brandstetter, Kapoor, Gupta, and Grover]{nguyen2023climax}
Tung Nguyen, Johannes Brandstetter, Ashish Kapoor, Jayesh~K. Gupta, and Aditya Grover.
\newblock Climax: {A} foundation model for weather and climate.
\newblock \emph{CoRR}, abs/2301.10343, 2023.
\newblock \doi{10.48550/ARXIV.2301.10343}.
\newblock URL \url{https://doi.org/10.48550/arXiv.2301.10343}.

\bibitem[Paischer et~al.(2025)Paischer, Galletti, Hornsby, Setinek, Zanisi, Carey, Pamela, and Brandstetter]{paischer2025gyroswin}
Fabian Paischer, Gianluca Galletti, William Hornsby, Paul Setinek, Lorenzo Zanisi, Naomi Carey, Stanislas Pamela, and Johannes Brandstetter.
\newblock Gyroswin: 5d surrogates for gyrokinetic plasma turbulence simulations, 2025.
\newblock URL \url{https://arxiv.org/abs/2510.07314}.

\bibitem[Pathak et~al.(2022)Pathak, Subramanian, Harrington, Raja, Chattopadhyay, Mardani, Kurth, Hall, Li, Azizzadenesheli, Hassanzadeh, Kashinath, and Anandkumar]{pathak2022fourcastnet}
Jaideep Pathak, Shashank Subramanian, Peter Harrington, Sanjeev Raja, Ashesh Chattopadhyay, Morteza Mardani, Thorsten Kurth, David Hall, Zongyi Li, Kamyar Azizzadenesheli, Pedram Hassanzadeh, Karthik Kashinath, and Animashree Anandkumar.
\newblock Fourcastnet: {A} global data-driven high-resolution weather model using adaptive fourier neural operators.
\newblock \emph{CoRR}, abs/2202.11214, 2022.
\newblock URL \url{https://arxiv.org/abs/2202.11214}.

\bibitem[Price et~al.(2025)Price, Sanchez{-}Gonzalez, Alet, Andersson, El{-}Kadi, Masters, Ewalds, Stott, Mohamed, Battaglia, Lam, and Willson]{sanchez2025gencast}
Ilan Price, Alvaro Sanchez{-}Gonzalez, Ferran Alet, Tom~R. Andersson, Andrew El{-}Kadi, Dominic Masters, Timo Ewalds, Jacklynn Stott, Shakir Mohamed, Peter~W. Battaglia, R{\'{e}}mi~R. Lam, and Matthew Willson.
\newblock Probabilistic weather forecasting with machine learning.
\newblock \emph{Nat.}, 637\penalty0 (8044):\penalty0 84--90, 2025.
\newblock \doi{10.1038/S41586-024-08252-9}.
\newblock URL \url{https://doi.org/10.1038/s41586-024-08252-9}.

\bibitem[Schlichting \& Gersten(2016)Schlichting and Gersten]{schlichting2016boundary}
Hermann Schlichting and Klaus Gersten.
\newblock \emph{Boundary-Layer Theory}.
\newblock Springer, Berlin, Heidelberg, 9 edition, 2016.
\newblock \doi{10.1007/978-3-662-52919-5}.

\bibitem[Schusterbauer et~al.(2024)Schusterbauer, Gui, Ma, Stracke, Baumann, Hu, and Ommer]{schusterbauer2024boosting}
Johannes Schusterbauer, Ming Gui, Pingchuan Ma, Nick Stracke, Stefan~A. Baumann, Vincent~Tao Hu, and Björn Ommer.
\newblock Boosting latent diffusion with flow matching.
\newblock In \emph{ECCV}, 2024.

\bibitem[Shi et~al.(2024)Shi, Ma, Ma, and Li]{shi2024scaling_timeseries}
Jingzhe Shi, Qinwei Ma, Huan Ma, and Lei Li.
\newblock Scaling law for time series forecasting.
\newblock In Amir Globersons, Lester Mackey, Danielle Belgrave, Angela Fan, Ulrich Paquet, Jakub~M. Tomczak, and Cheng Zhang (eds.), \emph{Advances in Neural Information Processing Systems 38: Annual Conference on Neural Information Processing Systems 2024, NeurIPS 2024, Vancouver, BC, Canada, December 10 - 15, 2024}, 2024.
\newblock URL \url{http://papers.nips.cc/paper\_files/paper/2024/hash/97c2f0fac182353062d304d0322ae285-Abstract-Conference.html}.

\bibitem[Song \& Tartakovsky(2021)Song and Tartakovsky]{song2021transfer_learning_mf}
Dong~H. Song and Daniel~M. Tartakovsky.
\newblock Transfer learning on multi-fidelity data.
\newblock \emph{CoRR}, abs/2105.00856, 2021.
\newblock URL \url{https://arxiv.org/abs/2105.00856}.

\bibitem[Subramanian et~al.(2023)Subramanian, Harrington, Keutzer, Bhimji, Morozov, Mahoney, and Gholami]{subramanian2023towards_foundation_models}
Shashank Subramanian, Peter Harrington, Kurt Keutzer, Wahid Bhimji, Dmitriy Morozov, Michael~W. Mahoney, and Amir Gholami.
\newblock Towards foundation models for scientific machine learning: Characterizing scaling and transfer behavior.
\newblock In Alice Oh, Tristan Naumann, Amir Globerson, Kate Saenko, Moritz Hardt, and Sergey Levine (eds.), \emph{Advances in Neural Information Processing Systems 36: Annual Conference on Neural Information Processing Systems 2023, NeurIPS 2023, New Orleans, LA, USA, December 10 - 16, 2023}, 2023.
\newblock URL \url{http://papers.nips.cc/paper\_files/paper/2023/hash/e15790966a4a9d85d688635c88ee6d8a-Abstract-Conference.html}.

\bibitem[Team(2024)]{grattafiori2024llama3herdmodels}
LLaMA-3 Team.
\newblock The llama 3 herd of models, 2024.
\newblock URL \url{https://arxiv.org/abs/2407.21783}.

\bibitem[Weller et~al.(1998)Weller, Tabor, Jasak, and Fureby]{weller1998openfoam}
H.~G. Weller, G.~Tabor, H.~Jasak, and C.~Fureby.
\newblock A tensorial approach to computational continuum mechanics using object-oriented techniques.
\newblock \emph{Computers in Physics}, 12\penalty0 (6):\penalty0 620--631, 1998.
\newblock \doi{10.1063/1.168744}.

\bibitem[Wu et~al.(2024)Wu, Luo, Wang, Wang, and Long]{wu2024transolver}
Haixu Wu, Huakun Luo, Haowen Wang, Jianmin Wang, and Mingsheng Long.
\newblock Transolver: {A} fast transformer solver for pdes on general geometries.
\newblock In \emph{Forty-first International Conference on Machine Learning, {ICML} 2024, Vienna, Austria, July 21-27, 2024}. OpenReview.net, 2024.
\newblock URL \url{https://openreview.net/forum?id=Ywl6pODXjB}.

\bibitem[Yang et~al.(2024)Yang, Hu, Zhou, Liu, Shi, Li, Li, Chen, Chen, Zeni, Horton, Pinsler, Fowler, Zügner, Xie, Smith, Sun, Wang, Kong, Liu, Hao, and Lu]{yang2024matter_sim}
Han Yang, Chenxi Hu, Yichi Zhou, Xixian Liu, Yu~Shi, Jielan Li, Guanzhi Li, Zekun Chen, Shuizhou Chen, Claudio Zeni, Matthew Horton, Robert Pinsler, Andrew Fowler, Daniel Zügner, Tian Xie, Jake Smith, Lixin Sun, Qian Wang, Lingyu Kong, Chang Liu, Hongxia Hao, and Ziheng Lu.
\newblock Mattersim: A deep learning atomistic model across elements, temperatures and pressures.
\newblock \emph{arXiv preprint arXiv:2405.04967}, 2024.

\bibitem[Yang \& Griffin(2021)Yang and Griffin]{yang2021grid}
Xiang I.~A. Yang and Kevin~P. Griffin.
\newblock Grid-point and time-step requirements for direct numerical simulation and large-eddy simulation.
\newblock \emph{Physics of Fluids}, 33\penalty0 (1), January 2021.
\newblock ISSN 1089-7666.
\newblock \doi{10.1063/5.0036515}.
\newblock URL \url{http://dx.doi.org/10.1063/5.0036515}.

\bibitem[Yang et~al.(2013)Yang, Koziel, and Leifsson]{yang2013simulation}
Xin-She Yang, Slawomir Koziel, and Leifur Leifsson.
\newblock Computational optimization, modelling and simulation: Recent trends and challenges.
\newblock \emph{Procedia Computer Science}, 18:\penalty0 855--860, 2013.
\newblock ISSN 1877-0509.
\newblock \doi{https://doi.org/10.1016/j.procs.2013.05.250}.
\newblock URL \url{https://www.sciencedirect.com/science/article/pii/S1877050913003931}.
\newblock 2013 International Conference on Computational Science.

\bibitem[Yao et~al.(2025)Yao, Yang, Jiang, Liang, Jin, and Pan]{yao2025scaling_timeseries}
Qingren Yao, Chao{-}Han~Huck Yang, Renhe Jiang, Yuxuan Liang, Ming Jin, and Shirui Pan.
\newblock Towards neural scaling laws for time series foundation models.
\newblock In \emph{The Thirteenth International Conference on Learning Representations, {ICLR} 2025, Singapore, April 24-28, 2025}. OpenReview.net, 2025.
\newblock URL \url{https://openreview.net/forum?id=uCqxDfLYrB}.

\bibitem[Zeni et~al.(2025)Zeni, Pinsler, Z{\"u}gner, Fowler, Horton, Fu, Wang, Shysheya, Crabb{\'e}, Ueda, et~al.]{zeni2025generative_materials}
Claudio Zeni, Robert Pinsler, Daniel Z{\"u}gner, Andrew Fowler, Matthew Horton, Xiang Fu, Zilong Wang, Aliaksandra Shysheya, Jonathan Crabb{\'e}, Shoko Ueda, et~al.
\newblock A generative model for inorganic materials design.
\newblock \emph{Nature}, pp.\  1--3, 2025.

\bibitem[Zhai et~al.(2022)Zhai, Kolesnikov, Houlsby, and Beyer]{zhai2022scaling_vit}
Xiaohua Zhai, Alexander Kolesnikov, Neil Houlsby, and Lucas Beyer.
\newblock Scaling vision transformers.
\newblock In \emph{{IEEE/CVF} Conference on Computer Vision and Pattern Recognition, {CVPR} 2022, New Orleans, LA, USA, June 18-24, 2022}, pp.\  1204--1213. {IEEE}, 2022.
\newblock \doi{10.1109/CVPR52688.2022.01179}.
\newblock URL \url{https://doi.org/10.1109/CVPR52688.2022.01179}.

\bibitem[Zhang et~al.(2024)Zhang, Zhang, Santoni, Khosronejad, and Samaras]{zhang2024windfarm}
Dichang Zhang, Zexia Zhang, Christian Santoni, Ali Khosronejad, and Dimitris Samaras.
\newblock Transfer learning in multi-fidelity surrogate modeling: A wind farm case.
\newblock In \emph{ICML 2024 AI for Science Workshop}, 2024.
\newblock URL \url{https://openreview.net/forum?id=yBTDCqNcan}.

\end{thebibliography}

\newpage
\appendix

\section*{Appendix}
\section{Training details}
\label{app:training_details}

We train our Transolver \citep{wu2024transolver} models using AdamW \citep{loshchilov2019adamw} with a weight decay of $1 \times 10^{-4}$, $\beta_1 = 0.9$ and $\beta_2 = 0.999$ for 500 epochs, applying early stopping if there is no improvement in validation loss for 250 consecutive epochs.
We employ a cosine decay learning rate scheduler with a 10 epoch linear warmup to an initial learning rate of $5\times 10^{-4}$.
We use gradient clipping and train in single precision float-32.
We list the exact hyperparameter choices contributing to the total model size of $\sim$4M params in \cref{tab:transolver_hyperparams}.

\begin{table}[h]
\centering
\caption{Transolver hyperparameters used.}
\label{tab:transolver_hyperparams}
    \begin{tabular}{lr}
    \toprule
    \textbf{Hyperparameter} & \textbf{Value} \\
    \midrule
    Base dimension                      & 256 \\
    \# Attention heads                  & 4 \\
    \# Transformer layers               & 8 \\
    Slice base                          & 128 \\
    MLP expansion ratio                 & 2 \\
    Dropout (MLPs/projections)          & 0.1 \\
    Dropout (Attention)                 & 0.1 \\
    \bottomrule
    \end{tabular}
\end{table}

\section{Normalized Mean Absolute Error}
\label{app:nmae}

We define the \ac{nMAE} as
\[
\mathrm{nMAE}
=
\frac{
    \sum_{i=1}^{N} \bigl| \hat{y}^{\mathrm{LF}}_{i} - y_{i}^{\mathrm{HF}} \bigr|
}{
    \sum_{i=1}^{N} \bigl| y_{i}^{\mathrm{HF}} \bigr|
},
\]

where $\hat{\boldsymbol{y}}_{i}^{\mathrm{LF}}$ are the fields from the low-fidelity simulation interpolated onto the corresponding high-fidelity mesh  (nearest neighbor) and $N$ is the number of mesh points of the high-fidelity sample.
In our comparison, we report the average \ac{nMAE} over all test samples.


\end{document}